\documentclass[11pt,a4paper]{article}
\usepackage{times,latexsym}
\usepackage{url}
\usepackage{enumitem}
\usepackage[T1]{fontenc}

\usepackage[acceptedWithA]{tacl2018v2}
\usepackage{graphicx}
\usepackage{hyperref}

\usepackage{xspace,mfirstuc,tabulary}

\renewcommand{\confidential}{}
\renewcommand{\anonsubtext}{}
\newcommand{\instr}

\iftaclpubformat 

\else

\fi

\title{Recent Advances in Neural Text Generation: A Task-Agnostic Survey}

\author{
 Chen Tang, Frank Guerin \and Chenghua Lin \\
 Department of Computer Science, University of Surrey, Guildford, United Kingdom
  \\
  Department of Computer Science, The University of Sheffield, Sheffield, United Kingdom \\
  {\sf \{chen.tang, f.guerin\}@surrey.ac.uk \and c.lin@sheffield.ac.uk} \\
}

\date{}
\begin{document}
\maketitle
\begin{abstract}
In recent years, considerable research has been dedicated to the application of neural models in the field of natural language generation (NLG). The primary objective is to generate text that is both linguistically natural and human-like, while also exerting control over the generation process. This paper offers a comprehensive and task-agnostic survey of the recent advancements in neural text generation. These advancements have been facilitated through a multitude of developments, which we categorize into four key areas: data construction, neural frameworks, training and inference strategies, and evaluation metrics. By examining these different aspects, we aim to provide a holistic overview of the progress made in the field. Furthermore, we explore the future directions for the advancement of neural text generation, which encompass the utilization of neural pipelines and the incorporation of background knowledge. These avenues present promising opportunities to further enhance the capabilities of NLG systems. Overall, this survey serves to consolidate the current state of the art in neural text generation and highlights potential avenues for future research and development in this dynamic field.
\end{abstract}
\section{Introduction}
Natural Language Generation (NLG) is a highly challenging sub-field of Natural Language Processing (NLP)~\cite{reiter_dale_2000}, which incorporates knowledge (e.g. text, pictures, audio, tables, etc.) and generates corresponding task-oriented text as output,  e.g. weather forecast reports. NLG has a range of applications~\cite{gatt2018survey,santhanam2019survey}.
NLG is traditionally tackled by symbolic and rule-based approaches, but in recent years  deep learning techniques have attracted a great amount of interest  \cite{belinkov2019analysis}.
Both approaches have their strengths, and it is possible that a neurosymbolic approach will dominate in future. This survey will focus on recent advances in the neural approach.

Existing survey papers usually summarize one of the NLG applications such as Story Generation \cite{hou2019survey,alhussain2021automatic,tang-etal-2022-ngep,huang-etal-2022-improving}, Text summarisation \cite{suleiman2020deep,el2021automatic}, Dialogue~\cite{ni2021recent,tang2022terminology,zhanga2023cadge}, Machine Translation~\cite{yang2020survey,dabre2020survey}, etc~\cite{loakman2023phonetically}. Only a few studies~\cite{lu2018neural,chandu2020positioning,jin2020recent,dong2021survey} discuss the development of the whole NLG area.

Task-specific surveys are beneficial, but the survey of  the whole NLG area could give broader ideas as inspiration for various applications using generative techniques, as the surveys like \citet{belinkov2019analysis,rogers2020primer} did. 
To our knowledge, this is the first comprehensive survey of neural text generation which summarises the commonalities and trends of recent advances.

From the perspective of neural text generation,  various NLG applications have task-agnostic commonalities: (\romannumeral1) Unlike traditional systems, neural networks capture features without ad-hoc feature engineering \cite{belinkov2019analysis}; (\romannumeral2) Various neural generative frameworks  mostly use similar encoder-decoder architectures, so they have common modules and training (or inference) strategies; (\romannumeral3) Evaluation metrics also have good generalization in NLG. Therefore, neural text generation has similar challenges \cite{chandu2020positioning,thomson2021generation} and solutions to analyze.

In this survey, we provide an overview of neural text generation via summarising the papers mainly published within the last 5 years\footnote{A tool is built to partly collect high-quality related samples: \href{https://github.com/tangg555/acl-anthology-helper}{https://github.com/tangg555/acl-anthology-helper}.}. We compartmentalize and analyze the contributions of these papers according to 4 aspects: \S\ref{sec:2} probes the characteristics of data construction for NLG tasks. \S\ref{sec:3} summarizes common deep learning techniques used in neural generative frameworks. In \S\ref{sec:4}, we analyze training and inference strategies of neural frameworks. \S\ref{sec:5} reviews and categorizes existing evaluation metrics for text generation. Following our analysis of recent advances, we discuss the future directions of research on Neural text generation including developing neural pipelines and exploiting background knowledge.

\section{Data Construction}
\label{sec:2}
 Datasets can be divided into benchmarks (for training and evaluating) and resources (usually external to tasks such as ConceptNet \cite{speer2017conceptnet}). A number of papers are devoted to setting up new datasets for NLG tasks, from which we investigate their demands and common approaches. 

\subsection{The Requirements of Data Construction}
\label{sec:2.1.1}
\subsubsection{Task-oriented Requirements}
New innovations in neural text generation often require a new dataset. For example, \citet{rashkin2020plotmachines} propose the task of outline-conditioned storytelling, which requires writing stories according to a given plot outline (more reasonable than given a leading context \cite{fan-etal-2018-hierarchical}), but existing datasets offer no plot outlines. Therefore, they present their datasets and corresponding construction workflow\footnote{Through RAKE (Rapid Automatic Keyword Extraction), which is a keyword extraction algorithm. \url{https://pypi.org/project/rake-nltk/}} satisfying the requirements. 

Similarily, other requirements e.g. modeling emotions \cite{huang-etal-2018-automatic,brahman-chaturvedi-2020-modeling} or persona \cite{chandu2019my,chan-etal-2019-modeling,wu-etal-2021-personalized} all need extra annotations for the new formalization of tasks. 

Some tasks also require high-level features (topics, themes, etc.). For example, to support new situations of chatbot detection \citep{gros2021rua}, contradiction detection \citep{nie2020like}, personal information leakage detection \citep{xu2020personal}, etc.,  Dialogue systems are trained on specific corpora to to deal with these task-oriented questions. 

Many new proposed tasks may even construct a novel dataset from scratch due to their special requirements which no existing ones satisfy  e.g. Personalized News Headline Generation \cite{ao2021pens}, Automatic Shellcode Generation \cite{liguori2021shellcode_ia32}, Controlled Table-To-Text Generation \cite{parikh2020totto}.

\subsubsection{Task-agnostic Requirements}
As available infrastructures for NLG, there are plenty of benchmarks \cite{rosenthal2017semeval, gardent2017webnlg, novikova2017e2e,wang2018glue} and data resources \cite{speer2017conceptnet, vrandevcic2014wikidata,clarke2001freenet, lehmann2015dbpedia} offering public resources, and related tools for evaluating, retrieving, constructing for NLG tasks. These are typically updated regularly. For instance, ConceptNet\footnote{\url{https://conceptnet.io/}} is a well-known semantic network widely used to pretrain neural networks to capture semantic features, with a first version in 2004 \cite{liu2004conceptnet}, and most recent version in 2017 \cite{speer2017conceptnet}. 

According to these updates, we conclude the trends for datasets development as follows:
 \begin{itemize}[noitemsep,nolistsep]
     \item \textbf{Larger and more comprehensive:} Replenish the corpora e.g. ontology, and facts, or Supplement new knowledge e.g. multilingual translation and commonsense knowledge.
     \item \textbf{More diverse:} Extra tasks and evaluation metrics.
     \item \textbf{Higher quality:} Address  existing problems e.g. ambiguities, lack of information.
 \end{itemize}

These trends also reflect the requirements of text generation. To acquire commonsense knowledge, CommonGen \cite{lin-etal-2020-commongen} is released to test the capability of generative commonsense reasoning for NLG systems. \citet{kumar2020clarq} present a large-scale ClarQ for Clarification Question Generation, clarifying ambiguities existing in current datasets to some extent. XGLUE \cite{liang2020xglue} has more diversified tasks labeled with more languages than GLUE. \citet{akoury2020storium} propose STORIUM, a platform\footnote{\url{https://storium.com/}} to produce and evaluate long-form stories with richer context than existing datasets.

\subsection{Approaches of Data Construction}
In this part, we discuss some common approaches of data construction which can cover most NLG tasks we reviewed.

\subsubsection{Creating datasets from Public Resources}
Processing public resources is the most common way to make a new dataset from scratch, as they are natural and abundant.
Plenty of Challenges \cite{rosenthal2017semeval, gardent2017webnlg, novikova2017e2e} and tasks \citep{agarwal2020knowledge, chen2020kgpt, wang2021wikigraphs}  sample plain text, such as descriptive text (e.g. wikipedia), dialogues (e.g. tweet), or structured data (e.g. wikidata \cite{vrandevcic2014wikidata})), etc. as the raw material. 

Crowdsourcing and automatic pipelines both are commonly used to collect and annotate  samples acquired from public resources. CommonGen\footnote{\href{https://gem-benchmark.com/data_cards/CommonGen}{\url{https://gem-benchmark.com/data_cards/CommonGen}}} \cite{lin-etal-2020-commongen} samples frequent concept-sets from existing caption corpora, and employs crowd workers to write corresponding sentences for referenced concept-sets. \citet{chen2020kgpt} sample  from Wikidump\footnote{\url{https://dumps.wikimedia.org/}} and WikiData, and collect data-text pairs through an automatic pipeline.

\subsubsection{Adding Extra Annotations to Existing Datasets}\label{extraannot}
Corpora used for text generation are often task-specific, with unique characteristics that make it difficult for them  to be shared by other tasks. e.g. Dialogues (datasets for Dialogues Systems) are incompatible with Machine Translation requirements (requiring multilingual corpora). However, similar tasks are able to partially share the datasets. For example, variant tasks of Story Generation  (requiring stories as datasets) can make datasets by adding extra annotations of plots \cite{fan-etal-2018-hierarchical}, emotions \cite{mostafazadeh2016corpus} or persona \cite{huang2016visual,shuster2018engaging} to existing datasets. This construction approach is more efficient and reliable than constructing from scratch.

\subsubsection{Transforming the Representation of Knowledge}
Parallel Corpora are a special kind of dataset. As knowledge has a variety of representations (e.g., different languages, or different formats like text, tables, images, etc.), some specific NLG tasks (e.g. Machine Translation or Data-to-text task) which transform one knowledge representation to another (e.g. triples to text) need to be trained on parallel corpora. 

\begin{figure}[htpb]
\centering
\includegraphics[scale=0.35]{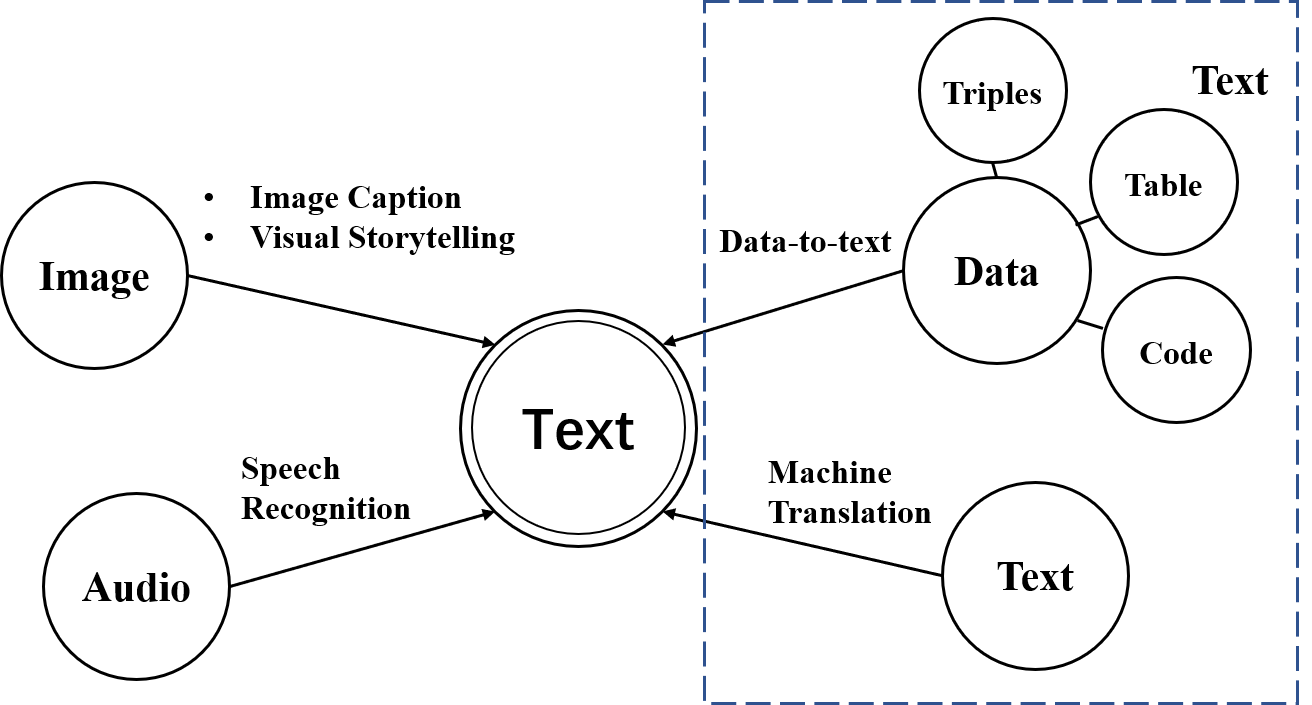}
\caption{Special NLG tasks require parallel corpora.}
\label{fig:parallel-tasks}
\end{figure}

These specific tasks are summarized in Figure \ref{fig:parallel-tasks}: Image Captions \cite{sharma2018conceptual,yoshikawa2017stair}, Visual Storytelling \cite{hsu-etal-2019-visual,chandu2019my}, Speech Recognition \cite{warden2018speech}, Data-to-text \cite{nishino2020reinforcement,fu2020partially,nishino2020reinforcement,richardson2017code2text}, Machine Translation \cite{hasan2020not,guzman2019flores}. In studies of these tasks, parallel corpora are indispensable \cite{belz2010extracting}.

Constructing parallel corpora is more challenging due to potential misalignment \cite{kuznetsova2013generalizing} and the fact that they do not generally occur naturally \cite{belz2010extracting}. The construction workflow is generally: crawling two types of knowledge, then aligning them manually or automatically, and finally filtering (discarding low-quality samples).

\section{Neural Frameworks}
\label{sec:3}
The methodologies of neural text can be generally viewed as a framework combining a range of deep learning techniques. There are various alternative deep learning techniques, and most of them can be composed together. However, there is no related study that demonstrated which alternative technique is superior (e.g. theoretically Autoregressive Decoding is more suitable for NLG tasks \cite{schmidt-etal-2019-autoregressive}, but sometimes well-designed Non-autoregressive ones could outperform their Autoregressive counterparts (\S\ref{non-autoregressive})). Thus, we separately analyze these techniques used to build a neural generative framework.

\subsection{Encoding Architectures} \label{position-encoding}
Encoding processes\footnote{\url{https://medium.com/analytics-vidhya/nlp-text-encoding-a-beginners-guide-fa332d715854}} embed input (text, or image/video, etc.) into continuous representation spaces for neural models, in which knowledge features are converted to the numeric vectors. With respect to text generation tasks, most encoding architectures are mainly based on three techniques: (\romannumeral1) Attention Mechanism e.g. Transformer \cite{vaswani2017attention}, (\romannumeral2) Variational Inference e.g. Seq2Seq Variational Autoencoder (Seq2Seq VAE) \cite{bowman-etal-2016-generating}, and (\romannumeral3) Adversarial Learning e.g. Sequence Generative Adversarial Nets (SeqGAN) \cite{yu2017seqgan}. To date, no obvious superiority has been demonstrated among these three types of encoder architectures.

Therefore, a number of studies have combined these three techniques, where the combinations had better performance. \citet{tolstikhin2018wasserstein} propose Wasserstein Auto-Encoders, which are the combination of (\romannumeral2) VAEs
and (\romannumeral3) GANs. \citet{bahuleyan-etal-2018-variational} propose a variational (\romannumeral1) attention mechanism
applied to (\romannumeral2) VAE encoder. \citet{zhang2019self} propose Self-attention Generative Adversarial Networks which combine (\romannumeral1) Attention Mechanism with (\romannumeral3) GANs.

Derivatives of (\romannumeral1) and (\romannumeral3) usually keep the vanilla encoder structure but implement different training and inference strategies to encode features. In comparison, encoders of (\romannumeral2) have flexible encoding architectures, depending on ad hoc choices for  probabilistic modeling, driven by task requirements. 

Another trend that emerges in several studies is to improve the encoding process by increasing the numbers of encoders which help to encode separate latent-variables, e.g. modeling personalization \cite{chan-etal-2019-modeling,wu-etal-2020-guiding}, discourse coherence \cite{guan-etal-2021-long,xu-etal-2021-discovering,yoo-etal-2020-variational}, and topology \cite{huang-etal-2020-knowledge,damonte-cohen-2019-structural}.

\subsection{Decoding Architectures} \label{decoding}
In general, the decoding of neural text generation refers to inferring the text output based on encoded features. Decoding architectures can be divided into (\romannumeral1) Autoregressive and (\romannumeral2)~Non-autoregressive. \cite{chandu2020positioning}

\textbf{Autoregressive Decoding} or Autoregressive Generation (AG) dynamically generates predictions in a recurrent manner\footnote{\url{https://www.eigenfoo.xyz/deep-autoregressive-models/} illustates how Autoregressive decoding works.}.  At time step $k$, the output $o_k$ corresponds to the input features $X_k$. At step $k+1$, the input features $X_{k+1}$ consist of $X_k$ and $o_k$.

This method's ability to model  sequential dependencies when decoding is  powerful for generation tasks \cite{schmidt-etal-2019-autoregressive}, but this characteristic also leads to problems \cite{schmidt-etal-2019-autoregressive,su-etal-2021-non} of inability to parallelise, relatively high latency in inference, and train-test discrepancies.

\textbf{Non-autoregressive Decoding}\label{non-autoregressive} or Non-autoregressive Generation (NAG) has faster inference speed \cite{guo2019non} than (\romannumeral1).  the output text sequence $O$ is directly generated from $X$. In comparison to (\romannumeral1),  there is no explicit correlation among outputs. Thus, Non-autoregressive Decoding process also results in two common problems: the fixed length of output and the conditional independence among predictions. These problems mean that existing generation quality of (\romannumeral2) lags behind  (\romannumeral1) \cite{su-etal-2021-non}.

A number of studies take both decoding methods into consideration in order to alleviate those problems. \citet{schmidt-etal-2019-autoregressive} introduce a conditional random field (CRF) \cite{sha2003shallow}  to a non-autoregressive baseline \cite{Schmidt2018DeepSS}, which ``\textit{expresses local correlations on the word level but keeps the state evolution  non-autoregressive}.'' Similar approaches \cite{ma-etal-2019-flowseq,zhang-etal-2020-fast,chi-etal-2021-align,qian-etal-2021-glancing} are proposed to deal with the independence assumptions of (\romannumeral2), which produce a better language model taking advantage of both decoding strategies.

\subsection{Reinforcement Learning}
Reinforcement Learning (RL) is learning decision-making based on discrete policies \cite{Arulkumaran2017}. Because of the characteristics of learning discrete policies, policy gradient methods are quite suitable to model indifferentiable rewards. Through well designed policy gradient methods \citep{guo2015generating,li-etal-2016-deep}, RL approaches train generative models towards indifferentiable rewards by minimizing the policy gradient loss \cite{chan-etal-2019-neural}.

Current trends show  rising interest in two directions: (a) Use RL for data augmentation \cite{nishino-etal-2020-reinforcement,liu-etal-2020-data,kedzie-mckeown-2019-good}; (b) Design a task-oriented \cite{yadav-etal-2021-reinforcement,yao-etal-2020-keep,liu-etal-2019-towards-comprehensive,zhou-etal-2019-unsupervised,cai-etal-2020-tag} or adaptive \cite{chan-etal-2019-neural} reward policy.

Considering the speciality of text generation tasks, in which evaluation metrics have no gold-standard automatic metric (see \S\ref{evaluation}), \citet{pasunuru2020dorb} showed how to automatically optimize multiple rewards.

\subsection{Data Retrieval}
In text generation, data retrieval is a technique to retrieve external material by key words. These supplementary data contain rich representations around the key words, helping to augment the training of generative language models.
Thus, they are leveraged to implicitly augment \cite{ijcai-song2018ensemble,hua-etal-2019-argument-generation,fabbri-etal-2020-template}, guide \cite{cai-etal-2019-skeleton,xu-etal-2020-megatron,gupta-etal-2021-controlling}, or generalize \cite{tigunova2020charm} the text generation process.  

Data retrieval methods were also used as the main framework in generating systems. For example, in the area of conversation systems, retrieval-based systems \cite{Isbell2000CobotIL,Ji2014AnIR} can search best-match text as the reply to a user-issued utterance. However, with the rise of deep learning, data retrieval is more often used as an auxiliary module for a neural generator, as mentioned in the previous paragraph. Thus, the recent studies of Data Retrieval  mainly focus on  supportive tools \cite{liu-etal-2020-data-centric}.

\subsection{Exploiting Sequential Features}
Text can be viewed as a sequence of characters, words, phrases, or other granules.
 The sequential features of text can be seen as the order among these granules. The phrase "Alice gave Bill" has same words as "Bill gave Alice", but they are totally different. 
 Related work can be divided into the following categories, according to the way in which they model sequential features.

\subsubsection{Recurrent Neural Frameworks}
Recurrent Neural networks (RNNs) have achieved great success in last couple of years. LSTM \cite{sundermeyer2012lstm}, GRU \cite{chung2014empirical}, HRED \cite{sordoni2015hierarchical}, VHRED \cite{serban2017hierarchical}, RIMs \cite{goyal2019recurrent} were developed  one after another. Furthermore, the mechanism of RNNs also deeply affects the design of current encoders \cite{Fabius2015VariationalRA,chien2019variational} and decoders \cite{radford2019language}. The aforementioned Variational Autoencoder (VAE), for instance, is modified to a Variational Recurrent Autoencoder (VRAE) for text generation  \cite{Fabius2015VariationalRA,chien2019variational}.

\subsubsection{Probabilistic Models}
Text sequence generation can be partly viewed as the estimation over conditional probability distributions (the real posterior distribution is intractable) of latent variables with given observations such as words \cite{zhang-etal-2018-probabilistic,liao-etal-2020-probabilistically}, sentences \cite{bahuleyan-etal-2019-stochastic}, etc. Linear-chain Conditional Random Field (CRF) \cite{cai-etal-2017-crf,gehrmann-etal-2019-improving}, for instance, is a typical Markov network to model dependencies among sequence tokens. Furthermore, if we set up variational inference with a neural network to approximate intractable posterior probability and train it with KL loss, this is the mechanism applied in VAE \cite{Kingma2014} and its variants \cite{bowman-etal-2016-generating,bahuleyan-etal-2018-variational,zhang2019improve}. 

\subsubsection{Hierarchical Stacked Framework}
The Hierarchical Recurrent Encoder-Decoder (HRED), 
introduced by \citet{sordoni2015hierarchical} for a Query Suggestion task, uses RNNs to encode both session-level and query-level features. The session-level RNN layer can be viewed to be hierarchically on top of the query-level RNN layer. It is a good way to model multi-granule representations in text, and has been improved on in numerous  subsequent works (e.g. VHRD \cite{zeng2019dirichlet} and PHVM \cite{shao-etal-2019-long}) of this hierarchical stacked structure in dialogues \cite{serban2016building,ren-etal-2019-scalable}, Data-to-text \cite{gong-etal-2019-table,shao-etal-2019-long}, etc. \cite{weber-etal-2018-hierarchical}.
      
\subsubsection{Memory Network}
Inspired by the memorizing mechanism of RNNs, \citet{weston2014memory} and \citet{sukhbaatar2015end} proposed the Memory Network, or Memory-argumented Neural Network, to store long-term memory. Instead of training a hidden state to store "memory" of the input sequence, Memory Networks introduce an external memory component which is jointly trained with inference. In this way, neural networks hold the uncompressed sequential knowledge when doing inference; by keeping a long-term memory it performs better. The Memory Network has been widely used to keep historical information for generation tasks \cite{zhang-etal-2017-flexible,fu-feng-2018-natural,chen2019working,tian2019learning,chen2021cross,wu-etal-2021-personalized}, especially those having a large number of inputs \cite{jang2019mnnfast}.

\subsubsection{Others}
As  BERT \cite{devlin-etal-2019-bert} with Transformer \cite{vaswani2017attention} blocks increased in popularity, the attention mechanism (see \S\ref{position-encoding}) has also become popular in recent years. Attention mechanisms implement position embeddings to encode sequential features, and encode position embeddings that can be divided into the absolute \cite{devlin-etal-2019-bert} and the relative \cite{shaw-etal-2018-self}. We recommend the overview of \citet{dufter2021position} which compares position encoding modules over 30 Transformer-like models to further study position embedding. In addition to position embedding, attention (self-attention) mechanisms can also observe the order between tokens by input embeddings. Thus, some researchers  leverage special tokens to encode the order between sentences and utterances~\cite{guan-etal-2021-long,li2021conversations,tang-etal-2022-etrica}.

\subsection{Leveraging Auxiliary Tasks}
 Auxiliary tasks can help enhance the representation learning ability of neural models. According to the relationship between auxiliary tasks and the main task, we can apply them in a serial (modules) or parallel (Multi-task Learning) way:

\begin{itemize}[noitemsep,nolistsep,leftmargin=*]
     \item \textbf{Multi-task Learning:} \label{multi-task-learning} Based on the hypothesis that correlated tasks have similar representation spaces, Multi-task Learning leverages heterogeneous training data to augment the representation learning ability of neural models. 

    \cite{zhou-etal-2019-multi,ide-kawahara-2021-multi,zhou-etal-2019-multi,li2021improving,xu2021agggen}.
    The main task shares part of training parameters with correlated auxiliary tasks to jointly learn how to encode features. \citet{li2021improving}, for instance, integrate number ranking (NR) and Importance Ranking (IR) with the data-to-text task to augment encoders when training embeddings. NR and IR are two auxiliary tasks, which help to extract more task related features from tables.

     \item \textbf{Modules in Frameworks:} Auxiliary tasks can also be set as the preliminary steps in neural frameworks. \citet{brahman2020modeling} introduce two Emotion-Reinforced models to track emotions of the protagonist for their generative model of storytelling. \citet{zhao-etal-2020-learning-simple} design two kinds of auxiliary tasks, "order recovery" and "masked content recovery", to capture sequential and semantic features for their multi-turn dialogue system.

 \end{itemize}

\subsection{Hybrid Frameworks} \label{hybrid}
It is well-known that neural networks are unstable and hard to accurately control, because their continuous vector spaces are uninterpretable to humans\footnote{Bad interpretability makes neural networks have several bottlenecks, and Neural Network interpretability is an emerging direction \cite{zhang2021survey,zhang2018visual}}. 
In comparison, traditional (non-neural) models are more interpretable but require expertise in feature engineering, whereas  neural models do not. Therefore, some studies propose hybrid frameworks to take advantage of both solutions.

We observe two trends behind recent studies in hybrid frameworks: (\romannumeral1) Improving existing traditional frameworks via partially replacing modules with neural networks, for example \citet{luo-etal-2020-make} build a neural text-stitch module to help their template-based data-to-text system reduce human involvement. Other examples include \citet{manning-2019-partially,gangadharaiah-narayanaswamy-2020-recursive}. (\romannumeral2) Introducing data-driven non-neural techniques to "guide" neural models. For example, \citet{wang-etal-2021-template} use template generated responses as extra input to neural dialogue systems. In a nutshell, (\romannumeral2) means that through offering higher-level control features (extra inputs to neural models), some auxiliary techniques such as  template-based methods  \cite{cao-etal-2018-retrieve,peng-etal-2019-text,wang-etal-2021-template}, rule-based methods \cite{yang-etal-2017-semi}, 
traditional frameworks \cite{zhai-etal-2019-hybrid,goldfarb2020content}, etc. can be introduced to guide neural models. \citet{zhai-etal-2019-hybrid}, for instance, introduce a symbolic "agenda generator" to perform text planning, and then use neural models to generate stories according to a produced agenda (script).

\section{Training and Inference Strategies}
\label{sec:4}
In addition to the framework architecture, good training and inference strategies are also  important to the performance of text generation. 

\subsection{Sampling Strategies}
During decoding  most generative models compare the likelihood (or unlikelihood \cite{welleck2019neural,li-etal-2020-dont,lagutin-etal-2021-implicit}) among sampled candidates to predict the next token. \citet{chandu2020positioning} summarise three sampling techniques according to the inference process: (\romannumeral1) Random Sampling, (\romannumeral2) Top-$k$ Sampling, and (\romannumeral3) Top-$p$ Sampling \cite{Holtzman2020The} (selecting a dynamic $k$ number of works according to a threshold probability value). The Top-$p$ sampling (\romannumeral3), also known as Nucleus Sampling, currently claim to be the best available decoding strategy.

Beyond decoding, sampling strategies also have an impact on the training process (i.e. Negative Sampling). Good negative samples have been demonstrated useful when training deep learning models \cite{goldberg2014word2vec,guan2020union,ghazarian2021plot}. Contrastive Learning (CL) \cite{cl-for-mt-acl21,simcls-acl21}, for instance, aims to
maximize that representation gap for machine translation between different languages of randomly collected negative samples.

Through distinguishing high-quality negative samples, neural models can better learn how to optimize related evaluation metrics. \citet{guan2020union} train a neural model to measure the overall quality of generated stories by sampling negative samples with commonly observed errors (Repetition, Substitution, Reordering, and Negation Alteration) in existing NLG models. \citet{ghazarian2021plot} hypothesize that heuristically generated negative samples are not adequate to reflect implausible texts' characteristics, so they construct plot-guided adversarial examples, and get better evaluation metrics trained on these samples.

\subsection{Masking Strategies} \label{mlm}
Masking  is a common tool widely used in language modeling both in encoding and decoding. For Non-autoregressive Decoding models (See \S\ref{decoding}) masks are used to pad future tokens before being received by the decoder. Note that we consider noising strategies such as token deletion, token replacing, etc as part of masking strategies, because they can also be viewed as "masks" put on the original input.

As for encoding, the encoder reads in a partly masked text sequence, which forces language modeling to learn representions conditioned on incomplete context. A typical use is Masked Language Models (MLM), which train models to restore missing parts (masked tokens) of input text. In order to resist perturbations with contextual features, neural generative models \cite{fedus2018maskgan,lewis-etal-2020-bart,liao-etal-2020-probabilistically,chen-etal-2020-distilling,pmlr-v97-song19d,ahmad-etal-2021-unified}, e.g. MaskGAN, MASS and BART, train models with an autoregressive decoder to infill the masked tokens in the original data with missing tokens, in which masks can be viewed as the noise to "overcome".

\subsection{Adversarial Training} \label{adversarial}
Adversarial Training is a kind of data augmentation strategy, which learns from adversarial samples, and improves the performance and robustness of neural models \cite{jiang2020robust,bai2021recent}. As a derivative, \citet{zhou-etal-2021-learning} propose Inverse Adversarial Training (IAT) which trains neural models to be sensitive to perturbations. It encourages neural dialogue systems to capture dialogue history, and avoid  generic responses via penalizing the same response under a perturbed dialogue history.

Adversarial training is able to address Adversarial Attacks \cite{wang-etal-2020-t3}, which refers to misguiding well-trained models through adding artificial perturbation. The  Generative Adversarial Network (GAN) is a typical technique of Adversarial Training, and promising in NLG area \cite{de2021survey}. GANs train both a generator to generate artificial samples resembling original ones, and a discriminator to resist perturbations on the data's distribution . 

However, as the vanilla GAN only suits continuous and differentiable output while in comparison text generation generates sequences of discrete tokens, many generative frameworks \cite{yu2017seqgan,zou-etal-2020-reinforced,zhou-etal-2021-learning} implement adversarial training through combining both GAN and Reinforcement Learning (RL) modules (e.g. the aforementioned SeqGAN encoder). There are also other ways \cite{cui-etal-2019-dal,wang-etal-2020-t3} to address discrete output for GAN models. \citet{wang-etal-2020-t3}, for instance, propose a tree-based auto-encoder embedding discrete tokens into a continuous space.

\subsection{Knowledge Distillation}
Knowledge Distillation \cite{Stanton2021DoesKD,Gou2021KnowledgeDA} is a very popular model compression and acceleration technique, which trains a student network to emulate a larger and cumbersome teacher model. In the work of \citet{chen-etal-2020-distilling}, they attempt to train an autoregressive model, taught by BERT \cite{devlin-etal-2019-bert}. \citet{haidar2019textkd} apply knowledge distillation to GANs for text generation.

To improve knowledge distillation on generative models, \citet{melas-kyriazi-etal-2019-generation} propose a training approach called "generation-disillation" which leverages data augmentation from 
an external Seq2Seq model to bridge the gap between "teacher" and "student". \citet{tang-etal-2019-natural} let the student network BiLSTM  emulate the distribution of input text instead of the conditional probability distribution of the teacher network.

\subsection{Pre-training} \label{pre-training}
Pre-training refers to training a neural language model with large-scale corpora to help it learn parameters, and this method is basically self-supervised and task-agnostic. In order to advance NLG tasks, the Pre-trained Language Models (PLMs) are fine-tuned in task-oriented frameworks with small changes to the workflow and retraining on the datasets.

PLMs are prevalent in current NLG tasks, because they alleviate the lack of training corpora and feature engineering problems \cite{qiu2020pre} for specific tasks. In light of taking advantage of Pre-training techniques, studies consider the improvement on both language models and high-quality corpora.

In the context of NLG, main-stream language models use the encoder-decoder architectures, and transfomers \cite{vaswani2017attention} have superseded RNNs \cite{peters-etal-2018-deep} as basic blocks to build pre-training neural layers. The aforementioned non-autoregressive language model, e.g. BERT \cite{devlin-etal-2019-bert} and autoregressive language model, e.g. GPT-2 \cite{radford2019language} both have been widely applied in downstream tasks as baselines or modules learning representations of knowledge. Current trends use training strategies such as mask sampling (See \S\ref{mlm}), multi-task learning \cite{guan-etal-2020-knowledge,xu-etal-2021-xlpt}, knowledge distillation \cite{gu-etal-2021-pral}, etc. to improve existing language models.

As for the pre-training corpora, some \cite{su2020moviechats,liang-etal-2020-xglue,zhang-etal-2020-dialogpt,xu-etal-2020-incorporating,qi-etal-2021-prophetnet} contribute to designing a construction workflow to build large-scale, high-quality, and task-oriented corpora. \citet{su2020moviechats}, for instance, construct a high-quality movie dialogue corpus, and experiments show that a simple neural approach trained on this corpus can outperform complex rule-based commercial systems.
Adjusting PLMs to heterogeneous knowledge (knowledge graph) \cite{chen2020kgpt_sec4,guan-etal-2020-knowledge,agarwal-etal-2021-knowledge} beyond text is another common approach to feed PLMs. \citet{guan-etal-2020-knowledge}, e.g., let GPT-2 learn commonsense knowledge on natural language sentences transformed by commonsense knowledge graphs.

\subsection{Other Methods}
There are some other strategies which are fundamental and common in NLG tasks. We list them as follows:
 \begin{itemize}[noitemsep,nolistsep,leftmargin=*]
     \item \textbf{Multi-task Learning (namely jointly learning).} Training multiple correlated neural tasks on shared network layers (introduced in Sec.~\ref{multi-task-learning}).
     \item \textbf{Teacher Forcing.} Training generative neural can be done in two ways: Free-running and Teacher Forcing. Current training stategies are usually based on these two methods, and we can view them as the variants \cite{lamb2016professor,goodman-etal-2020-teaforn,qi-etal-2020-prophetnet,zhang-etal-2019-bridging} of Teacher Forcing, e.g. Professor Forcing \cite{lamb2016professor}. They attempt to modify sampling strategies to address the Exposure Bias \cite{He2019QuantifyingEB}\footnote{Exposure Bias cause the gap of learned distribution between training and inference.} when using the Teacher Forcing strategy. 

     \item \textbf{Data Augmentation.} In general, data augmentation refers to every technique which increases the amount of training data, so we can see it in the aforementioned literature. e.g. Adversarial Training \citep{wang-bansal-2018-robust} creates negative samples to train, so this is also a kind of data augmentation.
 \end{itemize}
\section{Evaluation Metrics}\label{evaluation}
\label{sec:5}

Unlike other NLP tasks, NLG is open-ended, so usually NLG tasks have no gold-standard to measure the quality of generated text.
Human-centric metrics are unreliable and noisy \cite{otani-etal-2016-irt}, because they suffer from problems such as being expensive, time-consuming, individual biased and so on. Therefore, currently automatic metrics and human-centric metrics are both applied to evaluate NLG systems, and sometime extra correlation analysis \cite{guan-etal-2021-long,ghazarian2021plot} will be introduced between them e.g. Inter-Annotator Agreement (IAA) \cite{amidei-etal-2019-agreement}. Current NLG systems tend to choose multiple metrics among existing automatic and human-centric metrics to evaluate their results from multiple perspectives.

According to existing studies, we summarize current evaluation metrics (See examples in Table \ref{table:evaluation-examples}.) from two perspectives: the types of evaluators (humans, models or rules), and the data requirement (referenced, unreferenced, or hybrid). We will discuss the characteristics and trends in the following parts. 

\begin{table*}[t]
\small 
\centering
\begin{tabular}{p{1.8cm}p{4.8cm}p{3.8cm}p{4.0cm}}
\hline
& \bf Referenced & \bf Unreferenced & \bf Hybrid \\
\hline
\bf Untrained Automatic Metrics & N-gram based methods (heuristic rules) e.g. BLEU, ROUGE, METEOR, and MS-Jaccard. & Methods measuring Diversity e.g. Self-BLEU, Repetition, and Entropy.  & None observed \\
\hline
\bf Machine-Learned Metrics & Embedding based methods measuring similarity e.g. MEANT 2.0, MoverScore, and BertScore. & The Generator based methods or the Discriminator based methods. & Blended metheds e.g. RUBER, RUBER-BERT, and BLEAURT. \\
\hline
\bf Human-Centric Evaluation & Pairwise comparisons or Best–Worst Scaling. & Rating (or ranking) based methods. e.g. Likert Scale & Ensemble (or Aggregation) methods e.g. RankME. \\
\hline
\end{tabular}
\caption{\label{evaluation-examples} Evaluation metrics categorized according to "evaluators" e.g. Human-Centric Evaluation (humans give judgements), and "data requirement" e.g. Referenced (compare references with model generated text). Referenced metrics give a relative score through comparisons, while Unreferenced metrics will quantify NLG systems. 
  }
\label{table:evaluation-examples}
\end{table*}

\subsection{Untrained Automatic Metrics}
Untrained Automatic Metrics (sometimes abbreviated as Automatic Metrics) are classical methods for NLG tasks. They are mostly based on string handling techniques e.g. N-gram, word matching, string distance, etc. to measure the character-level distribution similarity with a reference text. 
N-gram based methods (e.g. BLEU \cite{papineni-etal-2002-bleu}, ROUGE \cite{lin-2004-rouge}, METEOR \cite{banerjee-lavie-2005-meteor}, and MS-Jaccard \cite{alihosseini-etal-2019-jointly}.) are the most typical referenced methods, which measure  similarity via overlaps. 

In comparison, unreferenced methods \cite{Zhu2018TexygenAB,yao2019plan} also calculate content overlaps but the reference is the model generated text. e.g. Self-BLEU \cite{Zhu2018TexygenAB} is proposed to measure diversity by calculating the average BLEU score between one sentence (as the hypothesis) and  others in a generated collection.

Untrained Automatic Metrics are biased, and poorly correlated to human judgements. \cite{reiter-2018-structured,sellam-etal-2020-bleurt}, but they are time-efficient and cheap, so they can be found in most NLG tasks to offer coarse evaluation from different perspectives \cite{novikova-etal-2017-need,celikyilmaz2020evaluation}.

\subsection{Machine-Learned Metrics}
Letting neural models learn to evaluate generated results is a promising method to overcome the disadvantages of Untrained Automatic Metrics: the coarse measurement and bad correlation to human judgements. When evaluating, machine-learning models map the measurement into a continuous space, so they can generate a more fine-grained real-valued score. In the training process, models learn how to generate or evaluate text as humans, which make the results have better correlation to human judgements.

Referenced Machine-Learned Metrics aim to measure the similarity between machine-generated text with either natural text or its counterparts. Therefore, they are mostly Embedding based methods (e.g. MEANT 2.0 \cite{lo-2017-meant}, MoverScore \cite{zhao-etal-2019-moverscore}, and BertScore \cite{zhang2020bertscore}), which compare the semantic representations by embedding \cite{clark-etal-2019-sentence}. Recent efforts are investigating how to generate more meaningful embeddings, for which we can get  some clues from the following: (\romannumeral1) word-level \cite{huang2016supervised} -> phrasal-level \cite{lo-2017-meant} -> sentence-level \cite{clark-etal-2019-sentence}; (\romannumeral2) fixture representations -> contextual representations \cite{zhang2020bertscore}; (\romannumeral3) coarse input -> fine input \cite{zhao-etal-2019-moverscore}. However, 
to measure similarity, the Referenced methods emphasize quality measurement while they neglect diversity\footnote{\url{https://textinspector.com/help/lexical-diversity/}} measurement.

Unreferenced Machine-Learned Metrics learn how to directly measure the output of generative models without any reference. Since most generative systems are developed to maximize the likehood of natural language \cite{yarats2018hierarchical}, the likelihood of human-written utterances generated by NLG systems (namely perplexity) can be used as the metrics. This kind of Generator based method \cite{ficler-goldberg-2017-controlling,yarats2018hierarchical,guan-huang-2020-union} is popular to compare a Neural model with its counterparts, but optimizing models only for perplexity may result in bland responses \cite{celikyilmaz2020evaluation}. On the other hand, Discriminator-based metrics (also known as the optimal discriminator) \cite{bowman-etal-2016-generating,kannan2017adversarial,hu2017toward} are trained through constructive learning to distinguish machine-generated text from human-written text. However, training the optimal discriminator needs large-scale human judgements or  overfitting may bring a strong bias \cite{garbacea-etal-2019-judge}. 

Recently, there is interest in building blended models (e.g. RUBER \cite{tao2018ruber}, RUBER-BERT \cite{ghazarian-etal-2019-better}, BLEAURT \cite{sellam-etal-2020-bleurt}), which design training strategies to enable end-to-end neural models to incorporate features belonging to the Referenced, Unreferenced or Untrained Automatic Metrics.

\subsection{Human-Centric Metrics}
Human-Centric Metrics are also called human judgements, which basically rely on the statistics of a crowdsourced collection (e.g. through Amazon Mechanical Turk). Human-centric metrics require the design of surveys. Human-Centric Metrics are important assessment criteria to evaluate NLG systems, but they are quite expensive and time-consuming. 

The Unreferenced human-centric metrics \cite{Galley2018EndtoEndCM,ghazarian2021plot} require the crowd workers to give a scaled rating score to the generated text. \citet{amidei-etal-2019-use} further studied 135 papers with Unreferenced human evaluation, and discuss good practices. On the other hand, the Referenced human-centric metrics \cite{li2019acute,clark2021choose,kiritchenko2017best,deriu2020spot}, which require crowd workers to give a relative rank for given systems (or natural language), are more complicated and domain-oriented (ad-hoc questions). This is due to  individual biases \cite{otani-etal-2016-irt} which make it hard to decide which system is better. Thus, some recent studies  \cite{novikova-etal-2018-rankme,sakaguchi-van-durme-2018-efficient,goldfarb2020content} also managed to combine both Referenced and Unreferenced human-centric metrics to establish an evaluation system.

\subsection{The Trends of Evaluation Metrics} \label{evaluation-trends}

\subsubsection{Combining Multiple Criteria}
We summarize the criteria used in the literature of NLG evaluation as follows, and we observe that newly designed evaluation methods tend to cover more metrics. 

\begin{itemize} [noitemsep,nolistsep,leftmargin=*]
    \item \textbf{Lexical:} Repetition, Distinctiveness, Perplexity.
    \item \textbf{Linguistic:} Fluency, Sentence Order, Naturalness, Discourse.
    \item \textbf{Semantic:} Coherence, Consistency, Relevance.
    \item \textbf{Overall:} Quality, Diversity, Informativeness, Preference.
\end{itemize}

\citet{guan-huang-2020-union,pang-etal-2020-towards,mehri-eskenazi-2020-usr,guan-etal-2021-openmeva,gehrmann-etal-2021-gem} are developing evaluation frameworks and benchmarks to combine multiple metrics.

\subsubsection{An End-to-end Evaluation Framework}
With the rapid development of deep learning techniques, Machine-Learned Metrics, especially End-to-end neural evaluation models, become a rapidly emerging research direction, because it is an easier way to combine multiple criteria. For example, previous metrics could evaluate either the quality or the diversity, while \citet{semeniuta2018accurate,hashimoto-etal-2019-unifying,guan-huang-2020-union,zhang-etal-2021-dynaeval} with a language model architecture, can capture both the quality and diversity of the generated samples.
 
\section{Discussion and Futher Directions} \label{sec:6}

\begin{figure*}[htpb]
\centering
\includegraphics[scale=0.5]{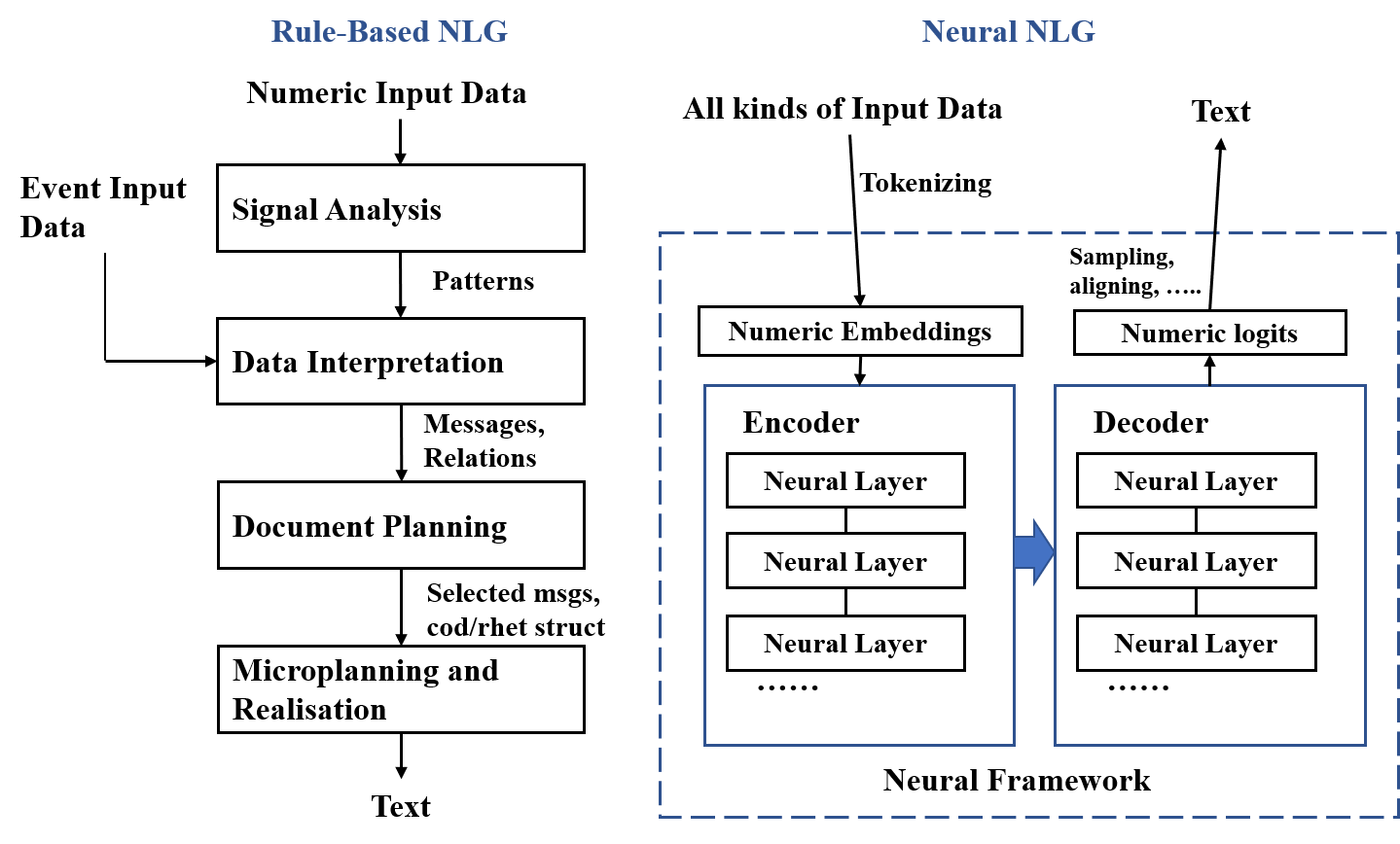}
\caption{Comparing the Data-to-text architectures between the rule-based system (left) and the neural generative system (right).}
\label{fig:frameworks}
\end{figure*}

\subsection{Development with Neural Pipeline Frameworks} 
Neural networks have largely changed the developing directions of generative systems. Traditional generative systems usually break up NLG tasks into separate stages that work as a pipeline \cite{reiter1997building}, e.g. the traditional rule-based system (taken from \citet{reiter2007architecture}) shown in Figure \ref{fig:frameworks} is a serial pipeline of four separate subdivided tasks: Signal Analysis, Data Interpretation, Document Planing, Microplanning and Realisation. These tasks are separate from each other, which means we could check the intermediate result under generation, and modify rules according to our expectations.

In comparison, neural frameworks are usually end-to-end, in which various modules or techniques (summarized in \S\ref{sec:3}) may take responsibility for separate functions such as encoding inputs \cite{radford2019language}, intents recognition  \cite{pinhanez-etal-2021-using}, emotions tracking \cite{brahman2020modeling}, content planning \cite{goldfarb2020content}, and so on. 

In Figure \ref{fig:frameworks}, we present a typical architecture of neural generative system (used in \citet{gong-etal-2019-enhanced,chang-etal-2021-neural,zhao-etal-2020-bridging,chen2020kgpt_sec4}). In a neural end-to-end framework, all kinds of inputs including target generated text are firstly mapped into numeric embeddings, and then neural modules feed forward information layer by layer. Finally the last output of the neural framework is used to generate the target tokens with a decoding strategy, and calculate the losses to optimize parameters. Modules in a framework connect to each other, and learn the inference principles through back propagation of losses between neural networks. This training and inference paradigm means that separate parts in a neural framework cannot be broken up independently.  It is hard to directly interfere in the intermediate operations in neural layers, because we do not understand what is in the hidden numeric space of embeddings. We can analyze the numeric embeddings only with indirect approaches, e.g., analyzing the meaning of self-attention outputs by drawing an ``attention'' heat map among input tokens \cite{li2020attention,cui-etal-2020-focus}.

On one hand, end-to-end neural frameworks are powerful. They take less manual effort and are widely accepted because of their good performance. On the other hand, end-to-end neural frameworks are not as flexible as traditional pipelines. The implicit features mapped in the numeric space make it harder to analyze and control text generation \footnote{Ehud Reiter who has made influential contributions in NLG area has discussed this recently (e.g. Challenges are Same for Neural and Rule NLG \url{https://ehudreiter.com/2021/11/08/challenges-same-neural-rule-nlg/})}. Therefore, end-to-end neural frameworks usually have problems of relatively bad generalization, bad robustness, and incomprehensible and uncontrollable working principles. 

As discussed above, one of the key limitations of the end-to-end paradigm is the uncontrollable working principles. In contrast, the traditional pipeline approaches e.g. heuristic rules, symbolic systems and template-based methods are more controllable and understandable. To address the key limitations of the end-to-end paradigm, several new strands of work have been proposed. One strand of work is developing neural pipeline approaches \cite{castro-ferreira-etal-2019-neural,nie-etal-2018-operation,moryossef-etal-2019-step,xu-etal-2020-megatron}, which introduce separate neural models in a traditional pipeline.  \citet{xu-etal-2020-megatron}, for instance, set up a pipeline of a Keywords Predictor, Knowledge Retriever, Contextual Knowledge Ranker, and Conditional Generator for story generation. The Keywords Predictor generates keywords for context planning, and then retrieves sentences with high rank from an external knowledge base. These are then sent to the Conditional Generator for story generation. The keywords and retrieved sentences are readable, and can be easily changed by the human designer (more examples seen in \S\ref{hybrid}).

Through experiments we can better analyze the advantage of neural pipeline approaches. \citet{castro-ferreira-etal-2019-neural} compared their neural pipeline approach for Data-to-text task with end-to-end counterparts, and draw a conclusion that in most tested circumstances, the pipeline approach generates more fluent and better-quality texts, especially for unseen domains. In addition to  worse generalization, end-to-end ones also have the problem of hallucination \cite{rohrbach-etal-2018-object} that generates text containing facts not given or not true in the inputs. Thus there have been increasing efforts in developing neural pipeline approaches which use neural models for better representation learning, and explicitly split the generation process for better controllability.

\subsection{Exploiting Background Knowledge}
One of the major challenges for NLG (and Artificial Intelligence more generally (see e.g. \citet{LakeBBSarxiv2016}) is to make effective use of background knowledge. Most NLG systems can train on a specific dataset, and can perform well in generating similar samples. However they have no other background knowledge to help them to deal with examples outside of the distribution. Ideally we would like a system to be like a human: when trained on a specific set of examples they can learn to deal with similar examples, but they also carry a huge amount of background knowledge which they can use to cope with examples outside of the distribution. The kinds of background knowledge that are useful include: 
general knowledge, conceptual knowledge, and knowledge for dealing with human emotions, personalities, etc.

One way in which this has been tackled is with a task-specific training, which has been summarized in \S\ref{sec:2.1.1}. For example, more annotations can be added to train neural models to learn how to track emotions, persona, and so on. \citet{sun-etal-2021-adding} add chit-chat to task-oriented datasets to train a chatbot. Experiments show the added chit-chat annotations lead to improved quality of dialogues.

A second approach employs neural framworks, especially pre-trained models (see \S\ref{pre-training}),
to acquire background knowledge from external knowledge bases. This has been used for example to acquire conceptual knowledge \cite{zhang-etal-2021-unsupervised}, commonsense knowledge \cite{guan-etal-2020-knowledge,lin-etal-2020-commongen,ji-etal-2020-language}, semantic knowledge \cite{ko-etal-2019-linguistically}, multimodal knowledge \cite{xing-etal-2021-km}, etc. 
These works use background knowledge to improve their models' performance on task-oriented requirements, but this knowledge could also be used across various NLG tasks.

Pre-training  can alleviate problems of bad generalization and bad robustness, and some advances have been made to address zero-shot (or few-shot) scenarios  \cite{chen2020kgpt_sec4,antoun-etal-2021-aragpt2}. However, the idea behind background knowledge acquisition is to be able to apply it flexibly in new varied scenarios. In this case, fine-tuning pre-trained models does not go far enough, because of the way the knowledge embodied in the models is tied to contexts in which it appears and is implicit. 
A more human-like background knowledge would need to be more explicit and freed from irrelevant contextual associations \citep[see e.g.][]{DBLP:journals/corr/abs-2002-06177}.

If we compare the artificial neural network (ANN) models and what we know about human processing we can see a contrast: the ANN models embody knowledge implicitly and do a very fast processing (for inference) to generate a result, which may be correct if the implicit knowledge has captured what is required for that particular query, or may be incorrect if not.
In contrast humans sometimes do a much slower processing requiring iterative inference \cite{VANBERGEN2020176}, which may require multiple forward and backward passes to reach a decision. In writing a text humans sometimes produce sentences rapidly, but at other times require deeper thought. The existence of  faster and slower processes seems to be a general feature of human cognition, and is  also evident in other cognitive tasks.
In the specific case of visual perception, research has shown that recognising an object in a difficult image (showing part of an object) can take up to two seconds, whereas a straightforward whole object image can be processed in 50ms \cite{benoni2020takes}.
\citeauthor{benoni2020takes} propose `a prolonged
iterative process combining bottom-up and top-down
components'.
\citeauthor{VANBERGEN2020176} note the advantages of recurrence in allowing a network to `adjust its computational depth to the task at hand'; i.e.
the number of such iterations is not fixed, and can take longer if the situation requires it; we can speculate that this is the case when dealing with more rare or `out of training distribution' samples.
In this process background knowledge can participate in the form of contextual knowledge which constrains expectations, or via generative models that embody prior knowledge about the world and can be used in a process of `analysis by synthesis' \cite{VANBERGEN2020176}.

From  Artificial Intelligence a similar message about the need for deeper or more complex processing emerges. For example \citet{davis2015commonsense}, in analysing the shortcomings of existing AI with respect to common sense, concluded that there is a need for alternative modes of reasoning; they state that ``commonsense reasoning involves many different forms of reasoning including
reasoning by analogy\ldots''. Analogical reasoning, whether for text or image processing, will require iterative inference including a top-down process \cite{guerin2021projection}.

\section{Conclusion}
Throughout the paper, we discussed the innovative paradigms behind recent advances from a task-agnostic perspective. From \S\ref{sec:2} to \S\ref{sec:5} we mainly summarized and analyzed the commonalities of neural text generation from 4 aspects: data construction, neural frameworks, training and inference stategies, and evaluation metrics. We compartmentalize different techniques or strategies people use to develop their generative systems. Some of them are continually affecting the whole NLG area (e.g. pre-training), and some are topics of much debate (e.g. which is the best encoding and decoding architecture for NLG tasks).

According to the analysis on these commonalities, we give some opinions about the development of neural frameworks from 2 directions in \S\ref{sec:6}: (\romannumeral1) More effort could be taken into exploring neural pipelines. Compared to end-to-end paradigm, neural pipelines take advantage of traditional techniques, and perform better with respect to generalization, interpretability, controbility, and robustness. (\romannumeral2) Many strands of work aim to make effective use of background knowledge. We highlight some good practices and promising directions to improve neural generative systems.

In this article, we attempt to offer a broad overview of how to create a better neural framework for text generation, and hope to offer some ideas to help solve challenges by building on the results shown by existing recent approaches. 


\section{Acknowledgement}
Many thanks to the reviewers who spend their time help us check and revise the paper. 
Thank all the people who gave us suggestions and responded to our talks. The authors also  gratefully acknowledge financial support from China Scholarship Council. (No. 202006120039)

\bibliography{custom}

\bibliographystyle{acl_natbib}

\end{document}